\documentclass[conference,compsoc]{IEEEtran}
\ifCLASSOPTIONcompsoc
  \usepackage[nocompress]{cite}
\else
  \usepackage{cite}
\fi

\ifCLASSINFOpdf
  \usepackage[pdftex]{graphicx}
  \graphicspath{{../pdf/}{../jpeg/}}
  \DeclareGraphicsExtensions{.pdf,.jpeg,.png}
\else
  \usepackage[dvips]{graphicx}
  \graphicspath{{../eps/}}
  \DeclareGraphicsExtensions{.eps}
\fi

\usepackage{amsmath}

\usepackage{amsthm}
\usepackage{amssymb}
\usepackage{amsfonts}
\usepackage[ruled,vlined]{algorithm2e}
\usepackage{algpseudocode}
\usepackage{array}

\usepackage{graphicx}
\usepackage{subfigure}

\usepackage{stfloats}

\usepackage{url}

\usepackage{booktabs}
\usepackage{multirow}
\usepackage{textcomp}
\usepackage{fontawesome}
\usepackage{xcolor}

\begin{document}
%
\title{LabObf: A Label Protection Scheme for Vertical Federated Learning Through Label Obfuscation}

\author{\IEEEauthorblockN{Ying He\IEEEauthorrefmark{1},
Mingyang Niu\IEEEauthorrefmark{1},
Jingyu Hua\IEEEauthorrefmark{1}, 
Yunlong Mao\IEEEauthorrefmark{1},
Xu Huang\IEEEauthorrefmark{2},
Chen Li\IEEEauthorrefmark{2} and
Sheng Zhong\IEEEauthorrefmark{1}}
\IEEEauthorblockA{\IEEEauthorrefmark{1}dept. Computer and Science Technology, Nanjing University\\
Nanjing, China}
\IEEEauthorblockA{\IEEEauthorrefmark{2}Beijing Science and Technology Co, Three Fast Online\\
Beijing, China}}

\maketitle

\begin{abstract}
Split Neural Network, as one of the most common architectures used in vertical federated learning, is popular in industry due to its privacy-preserving characteristics. In this architecture, the party holding the labels seeks cooperation from other parties to improve model performance due to insufficient feature data. Each of these participants has a self-defined bottom model to learn hidden representations from its own feature data and uploads the embedding vectors to the top model held by the label holder for final predictions. This design allows participants to conduct joint training without directly exchanging data. However, existing research points out that malicious participants may still infer label information from the uploaded embeddings, leading to privacy leakage. In this paper, we first propose an embedding extension attack manipulating embeddings to undermine existing defense strategies, which rely on constraining the correlation between the embeddings uploaded by participants and the labels. Subsequently, we propose a new label obfuscation defense strategy, called `LabObf', which randomly maps each original integer-valued label to multiple real-valued soft labels with values intertwined, significantly increasing the difficulty for attackers to infer the labels. We conduct experiments on four different types of datasets, and the results show that LabObf significantly reduces the attacker's success rate compared to raw models while maintaining desirable model accuracy.
\end{abstract}


%
\IEEEpeerreviewmaketitle

\section{Introduction}
In the era of data deluge, the resources held by individuals or agencies, whether in terms of hardware or software, are gradually proved to be inadequate to handle the exponentially growing volume of data and the extensive task of mining potential user demands. Vertical federated learning (VFL) has emerged as a multi-party collaborative framework in response to this challenge. Compared to traditional horizontal federated learning, VFL is more suitable for scenarios where participants hold data with different features and do not want to share them directly. It finds extensive applications in fields such as healthcare\cite{ihvfl}, finance\cite{webank}, and Internet of Things (IoT)\cite{IoTvfl}.

Split Neural Network\cite{splitnn} is one of the most widely used organizational structures in real-world VFL scenarios. In this setup, the dataset is vertically divided into several parts, each held by different participants (usually two parties), with label information held exclusively by one participant. The participant holding the labels is typically called \textit{the label party} or \textit{the host}, while the other participants are referred to as \textit{clients}. The host is typically the initiator of the federated task, seeking to improve the primary task's performance by aggregating data from other entities (i.e., clients). Generally, the host has both a bottom model and a top model, while each client holds its own bottom model. The participants use their bottom models to learn embeddings from their own feature data and then upload these embeddings to the host. The host concatenates the embeddings into a single input for the top model, which then outputs the final prediction. The layer used for interaction between the bottom models and the top model is called \textit{the cut layer}. This structure is adopted by several renowned companies and communities, such as ByteDance\cite{bytedance}, Tencent\cite{tencent}, and OpenMined\cite{openmined}, for two-party VFL.

Although such architecture has been proven to possess strong privacy protection capabilities\cite{splitsec1,splitsec2}, existing research indicates that they still suffer from security vulnerabilities\cite{shortestpathattack,batchattack,labelinferattack,featureinfer}. Data owned by participants, especially label data crucial for prediction tasks, is at risk of being stolen. These risks exist in the information exchange process at the cut layer. For instance, malicious clients may steal the labels by exploiting the back-propagation of gradients from the top model\cite{shortestpathattack,batchattack} or the distribution of embeddings in the cut layer\cite{labelinferattack}. The former methods, such as the shortest path attack\cite{shortestpathattack}, entails inferring label information by assessing the distance between back-propagated gradients of target samples and those from the $0$-class and $1$-class samples. However, these attacks are primarily utilized in binary classification tasks and there are relatively mature solutions available to defend against them\cite{shortestpathattack,batchattack}. Therefore, this type of attack will not be discussed further in this paper.

The latter methods, which is also our main consideration, represented by the label inference attack\cite{labelinferattack}, capitalizes on the strong correlation between the embedding vectors from the local bottom model and labels. Attackers try to gather a small set of labeled samples to mimic training the top model through semi-supervised learning. Using the local `top model', they can then deduce label information from the intermediate data transmitted through the cut layer with high confidence. This method is highly adaptable, poses a significant threat to VFL models, and is currently a prevalent attack strategy. Moreover, in real-world scenarios, clients often possess the majority, or even all, of the attributes. In such cases, this attack would be more effective and exhibit greater aggressiveness.

In response to this kind of label leakage, some works attempt to protect labels through perturbation \cite{marvell,fedpass,flsg}. These methods essentially add fine-grained noise to the embeddings or back-propagated gradients to blur the differences between embeddings of different classes without affecting the main task's accuracy, thereby misleading or disrupting label inference attacks. This approach is advantageous because it is lightweight and does not require modifications to the main task, yielding good results when the attacker’s capabilities are limited (e.g., with an auxiliary set of fewer than $100$ samples). \textbf{However, such noise-adding schemes inevitably creates conflicts with the main task training due to its underlying principles.} Therefore, to balance data security and model accuracy, the magnitude of the noise must be limited. Consequently, when attackers have higher capabilities, such as a slightly larger auxiliary set of labeled samples (e.g., $500$ samples, which is still within the attacker’s capability but greatly exceeds the scope considered by such schemes), these schemes tend to lack robust defense.

Another representative defense approach is Discorloss\cite{discorloss}, which introduces an additional term in the loss function to constrain the correlation between the client's local embeddings and the labels, thereby increasing the difficulty for malicious clients to infer labels. This approach leverages the inherent redundancy of neural networks to help find a balance between main task accuracy and defense performance. However, this method primarily suppresses the direct distance correlation between the client's embeddings and labels, \textbf{which might be disrupted by the  malicious client as the corresponding embeddings are fully under her/his control.}

To further expose the above security vulnerabilities, we first introduce an attack called the Embedding Extension Attack to validate the flaw of Discorloss, demonstrating that it is risky to deploy defense by leveraging the client. Given that Discorloss relies on embeddings uploaded by the client for its calculations, the malicious client can add several perturbation dimensions to its output embedding. By assigning carefully crafted special values to these perturbation dimensions, the client can maintain a high correlation between the original embedding and the label while reducing the correlation between the embedding uploaded to the host and the label. This undermines Discorloss's constraint on the original embedding. Thus, as an attacker, it can still use the original embeddings that have evaded the constraints to infer labels.

Next, we propose a label obfuscation defense mechanism called \textit{LabObf}, which is highly effective and less susceptible to attacker interference, with the aim of specifically safeguarding label data from leakage. In this scheme, the host, as the defending party, secretly remaps the original labels to a new distribution space in a one-to-multiple manner. For instance, in the case of a binary classification dataset, the original $0$-class and $1$-class labels are each mapped to four real-valued soft labels. These eight soft labels are then scattered and exhibit a cross-distribution state in their values (for example, $0 \to [0.1, 0.3, 0.6, 0.9]$, $1 \to [0.2, 0.4, 0.5, 0.8]$). It is important to note that this process occurs on the host side, unknown to the client, and for each original label class, the mapping to each soft label is uniform random. This design aims to obscure the distribution of original class features at the cut layer through label obfuscation, making it harder for attackers to infer label information from their local embeddings. However, this approach presents a challenge in that training the main task of the model becomes more difficult, as the original data features lack a direct mapping to the soft labels. To ensure that the model learns this artificial mapping, modifications need to be made to the dataset, such as artificially constructing additional features to establish the mapping from feature data to soft labels. During this process, the host and client each hold a portion of additional features, and both parts jointly determine the mapping relationship to prevent the client from easily modifying or inferring the mapping.

\textbf{LabObf addresses some of the shortcomings of previous defense methods.} On the one hand, this scheme partially mitigates the conflict between primary task accuracy and defense strength that perturbation-based defense approaches often face. In this approach, the higher the training accuracy with soft labels, the higher the accuracy of the converted real labels, without compromising defense performance. This is because the defense strength is determined by the degree of obfuscation in the mapping relationship of the soft labels. On the other hand, the mapping between original labels and soft labels is controlled by the label holder, making it challenging for attackers to access or disrupt this process. Additionally, this approach has several highlights stemming from its own design. By training with the additional features and soft labels, the approach adds a layer of protection to reduce the direct information about real labels that could be inferred from the model parameters. Finally, such a scheme is practical and straightforward to implement in real-world scenarios.

Our contributions can be summarized as follows:
\begin{itemize}
    \item Validate the security of existing defense works: We propose a method called the Embedding Extension Attack, which adds perturbation dimensions to the embedding to help the label inference attack\cite{labelinferattack} bypass current defense solutions.
    \item Propose a robust defense mechanism against interference: We propose a method called LabObf, which maps original labels to a newly obfuscated distribution, reducing the risk of label information leakage at the cut layer.
    \item Conduct detailed empirical validation: We conduct experiments on several popular datasets including Epsilon\cite{epsilon_dataset}, Bank\cite{bank_marketing_222}, CovType\cite{covertype_31}, and Fault Type\cite{fault_dataset}. Among these, Epsilon and Bank are binary classification datasets, while CovType and Fault Type are multi-class classification datasets. The experimental results demonstrate that our proposed approach exhibits strong defense capabilities. LabObf can significantly lower the attack success rate to below that of scenarios where the bottom model does not leak any information, while maintaining the primary task accuracy within an acceptable range.
\end{itemize}

In the following sections, we describe VFL and the existing background knowledge of attacks and defenses in Sec.\ref{bg}. The threat model is presented in Sec.\ref{tm}. We first present an attack to demonstrate the flaw in the existing defense scheme in Sec.\ref{attack}, and then propose a new, more robust protection scheme in Sec.\ref{defense}. Sec.\ref{eval} and Sec.\ref{analysis} present experimental data and analysis, while Sec.\ref{relwork} analyzes related works. Finally, we conclude with a brief summary in Sec.\ref{conclu}.


\section{Background}\label{bg}
In this section, we first introduce the most commonly used VFL architecture we are focusing on - the split neural network (SplitNN). Then, we provide a brief overview of the existing label leakage risks faced by VFL. Finally, we introduce the mainstream label protection schemes.

\subsection{SplitNN-Based VFL}
VFL\cite{vfl} aims to enable secure collaboration among multiple parties holding data with different feature dimensions without directly exchanging the data. SplitNN\cite{splitnn} is one of the most common architectures designed to achieve this goal.

    \begin{figure}[htbp]
        \centering
        \includegraphics[width=\linewidth]{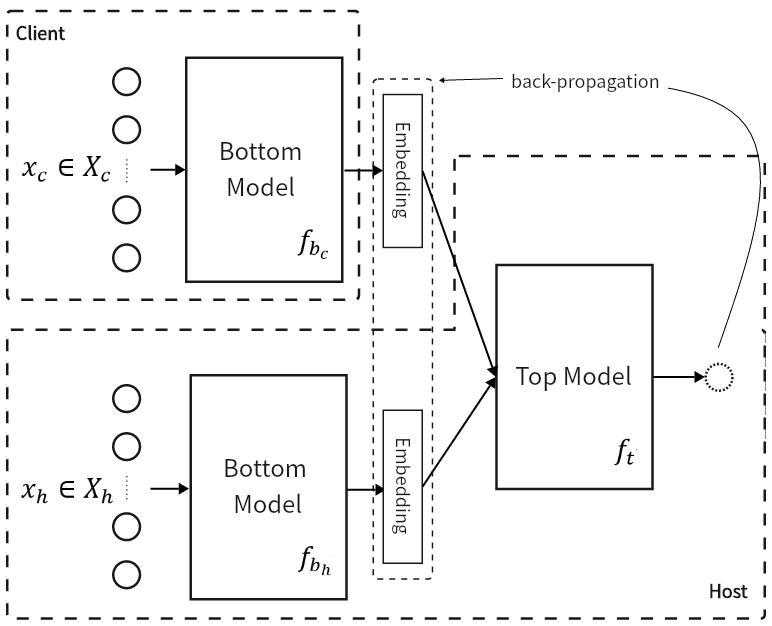}
        \caption{The architecture of SplitNN-based VFL}
        \label{fig:architecture}
    \end{figure}

Fig.\ref{fig:architecture} illustrates the architecture of a SplitNN setup with two parties. In terms of role allocation, the party that initiates the federated learning is called the host, typically the sole owner of the label data and having authority over the main task. The other party, responsible for providing data and auxiliary functions, is called the client. Regarding the dataset ($D = (X, Y)$) division, the features $X$ are divided into two parts, $X_c$ and $X_h$, with the former held by the client and the latter, along with the label data $Y$, held by the host. For network partitioning, the entire model is divided into two bottom models $f_{b_i}:\mathcal{X}_i\to \mathbb{R}^{d}$ ($i\in\{h,c\}$), and one top model $f_t:(\mathbb{R}^{d},\mathbb{R}^{d})\to\mathcal{Y}$. The client owns $f_{b_c}$, while the host owns $f_{b_h}$ and $f_t$. The host and client use their respective bottom models to learn the hidden representations of their respective feature data, which we call embeddings, and then upload these embeddings to the host. The host concatenates them and feeds the result into the top model for the main task's prediction. The entire model's objective function is shown in Eq.\ref{eq:splitnn}. During the back-propagation phase, the host sends the gradient of the cut layer to the bottom models, and each bottom model uses this gradient to compute and update its parameters. In the prediction phase, the label for sample $x = (x_c, x_h)$ is predicted collaboratively by the host and client, resulting in $\hat{y} = f_t(f_{b_c}(x_c),f_{b_h}(x_h))$.

    \begin{equation}\label{eq:splitnn}
        \mathcal{L}_c = \sum_{(x,y) \in D}\mathcal{L}_{ce}(f_t([f_{b_c}(x_c),f_{b_h}(x_h)]),y)
    \end{equation}
    
Although SplitNN can be applied to settings with multiple parties, in practice, VFL is often limited to two parties\cite{shortestpathattack,labelinferattack,marvell,discorloss,b18} due to issues of trust and efficiency. Therefore, this paper focuses exclusively on scenarios with two-party participation.

\subsection{Label Leakage Risks}
In vertical federated learning, label leakage typically arises from two data leakage sources: the first attempts to infer labels from the back-propagated gradients, while the second aims to deduce labels from the feature embeddings at the cut layer.

In the first type of attack, the shortest distance attack\cite{shortestpathattack} demonstrates label inference in binary classification tasks under VFL. It leverages back-propagated gradient information to compute the gradients to label $0$ and label $1$. It then compares which of these computed gradients is closer to the actual gradient: if the actual gradient is closer to the gradient for label $0$, the inferred label is $0$; otherwise, it's $1$. This is because, in binary classification, the back-propagated gradients tend to carry directional biases, thereby revealing label information. Batch label inference attack\cite{batchattack} targets label inference in multi-class classification problems with a softmax function layer. This method, through mathematical derivations, uses rank-related theorems to deduce label information based on the inherent mathematical operations of the softmax function. Although this approach has a robust mathematical basis for guaranteeing its attack efficacy, it is limited to cases where the top network layer consists of only a single softmax activation layer. When the architecture of the top network is unknown or additional fully connected layers are introduced, this approach becomes ineffective.

In the second type of attack, label inference attack\cite{labelinferattack} is the most typical method. It assumes that the malicious participant can access a very limited amount of label information, denoted as $D_{aux}$. The attacker uses a semi-supervised approach to train a local model $M_{shadow}$ that simulates the top model's classification process, and then fine-tunes the local model's results with the auxiliary set $D_{aux}$. The objective $\mathcal{L}_a$ of $M_{shadow}$ is calculated as follows:
    \begin{equation}
        \mathcal{L}_a = \sum_{(x,y) \in D_{aux}}\mathcal{L}_{ce}(M_{shadow}(f_{b_c}(x_c)),y)
    \end{equation}
where $\mathcal{L}_{ce}$ represents the cross-entropy loss function. By doing this, the attacker can infer labels for any sample it is interested in. This approach is effective because VFL optimizes the bottom models in such a way that their outputs provide more indicative features to support the top model's label predictions.

\subsection{Label Protection Schemes}
For the first type of attack, there are numerous existing defense measures that are relatively well-developed. Regarding the shortest distance attack, \cite{shortestpathattack} suggests adding noise in the direction of $g_0 - g_1$ to the backward gradients to counteract the directional bias. As for the batch label inference attack, \cite{batchattack} introduces a label disguise technique based on auto-encoder that transforms the label information into soft labels. In this scheme, the probabilities of each possible category in the soft label are elevated, reducing the gap between them and the target label's classification probability. The obfuscation strength of this scheme is relatively light and limited, as it does not forcibly break the boundaries between different categories but only seeks to blur the edges as much as possible. Essentially, it works by narrowing the prediction probability differences among different classes based on swapping labels. Consequently, the outputs from the bottom model still retain strong indicative features pointing towards the true classes, potentially exposing them to attacks from the second type of scheme.

For the second type of attack, there are several perturbation-based approaches that attempt to protect labels. Marvell\cite{marvell} is a noise-based label protection scheme designed for binary classification tasks. This scheme observes that in binary classification, the gradients for the two classes often differ in distribution. Based on the differences in the magnitude and direction of the gradients for different classes, this scheme selects optimal noise levels to balance model performance and label protection while limiting the noise magnitude. It should be noted that this scheme is tailored specifically for binary classification. FedPass\cite{fedpass} uses normal distribution sampling along with model parameters to design a more refined adaptive training noise, adjusting the variance of the normal distribution to control defense strength. The larger the variance, the greater the defense against label inference attacks, but this also leads to a corresponding decrease in model accuracy. FLSG\cite{flsg} similarly leverages gradient expansion, using randomly generated Gaussian-distributed gradients that resemble the original gradients, providing effective defense with minimal overhead. As mentioned earlier, these approaches add noise to increase the variance among samples within the same class and blur the distribution boundaries between different classes, aiming to strike a balance between primary task accuracy and disruption of attacks. However, these approaches do not fundamentally break the boundaries between the original classes. Consequently, when attackers hold attributes critical for classification or have access to a slightly larger auxiliary dataset, they can still infer label information from the embeddings. This is why this paper does not focus primarily on these approaches.

A representative defense scheme is Discorloss\cite{discorloss}. It introduces a new training objective $\mathcal{L}_d$ alongside the existing classification goal $\mathcal{L}_c$ for the top model. This new objective aims to minimize the distance correlation between the distribution of the cut layer's feature vectors and the label distribution. The specific objective functions are as follows:
        \begin{equation}
            \mathcal{L}_d = \sum_{(x,y) \in D}Discor(f_{b_c}(x_c),y)
        \end{equation}
        \begin{equation}\label{eq:discorloss}
            \mathcal{L} = \mathcal{L}_c + \lambda \mathcal{L}_d
        \end{equation}
where $Discor(\cdot)$ is the distance correlation function and $\lambda$ is an adjustable parameter used to balance the main task accuracy with the defense strength. By adopting this approach, the output from the bottom model becomes more dispersed, thereby reducing the likelihood of information leakage. However, it should be noted that relying on calculations based on data uploaded by potentially untrustworthy participants carries inherent risks.

In summary, the first type of attack scheme currently lacks generality, and there are relatively comprehensive defense measures that lead to low-security risks. The second type of attacks, however, is more scalable and aggressive, and we must point out that existing defense measures have limitations (as detailed in Sec.\ref{attack}), resulting in these attack schemes posing a significant threat to label information. Therefore, in Sec.\ref{defense}, we propose a more robust defense scheme to address this issue.

\section{Threat Model}\label{tm}
\subsection{Network Architecture}
Due to the widespread application of VFL tasks in practical environments involving two-party participation\cite{shortestpathattack,labelinferattack,marvell,discorloss,b18}, this paper primarily focuses on the SplitNN architecture involving two participating parties, namely the host and the client, like Fig.\ref{fig:architecture}. In this architecture, the client, potentially acting as an attacker, holds partial dimensional features and maintains the corresponding bottom model locally. During training and prediction, it is responsible for uploading locally computed embeddings to the host. The host, acting as the defending party, holds partial dimensional features and all labels, and maintains both the corresponding bottom and top models locally. During training and prediction, it calculates the corresponding embeddings locally and combines them with the embeddings uploaded by the client, and then inputs them into the top model. The output of the top model represents the prediction result.

\subsection{Attacker Modeling}
\textbf{Capabilities:}
\begin{itemize}
    \item The same as in \cite{shortestpathattack,labelinferattack,batchattack}, assuming that the attacker can obtain a small amount of labeled sample data. It is reasonable for the attacker, as a participant, to obtain a small amount of label information through offline purchases or theft.
    \item The attacker is unable to obtain or access the top model in any form.
    \item The attacker cannot obtain the predictions of the top model.
    \item The attacker can independently determine the structural characteristics of the local model it possesses.
    \item The attacker is not expected to damage the primary task's accuracy, as poor accuracy could cause the host to cease collaboration, leading to a loss of business and related benefits.
\end{itemize}

\textbf{Goal:} The attacker aims to infer sample label information from her/his local embeddings without compromising the primary task, using their known information and capabilities. Here, we assume the attacker employs the most threatening scheme currently known -- label inference attack\cite{labelinferattack}.

\subsection{Defender Modeling}
\textbf{Capabilities:}
\begin{itemize}
    \item The defender cannot access or obtain the bottom model held by the client in any form.
    \item The defender cannot obtain the feature data held by the client.
    \item The defender can independently determine the structural characteristics of the corresponding bottom model and the top model it possesses.
    \item The defender can independently determine the training objectives of the VFL model.
\end{itemize}

\textbf{Goal:} The defender's objective is to minimize the attacker's inference accuracy as much as possible, thereby reducing label information leakage, while ensuring the primary task's accuracy remains within an acceptable range.

\section{Embedding Extension Attack against Discorloss}\label{attack}
As mentioned in Sec.\ref{bg}, existing defense mechanisms represent and measure the relationship between embeddings and label information through a specific representation method. By reducing the result in this representation dimension, they lower the label information content in the embeddings, making it difficult for attackers to infer label information from a small amount of labeled data while allowing the VFL model to maintain its accuracy on the full dataset.

However, we must point out that this approach has inherent flaws. The calculation results are determined by the embeddings uploaded by the attacker. This means that the attacker can manipulate the embedding vectors to alter the correlation calculation results, effectively invalidating the defense mechanism.

\begin{figure}[htbp]
        \centering
        \includegraphics[width=0.9\linewidth]{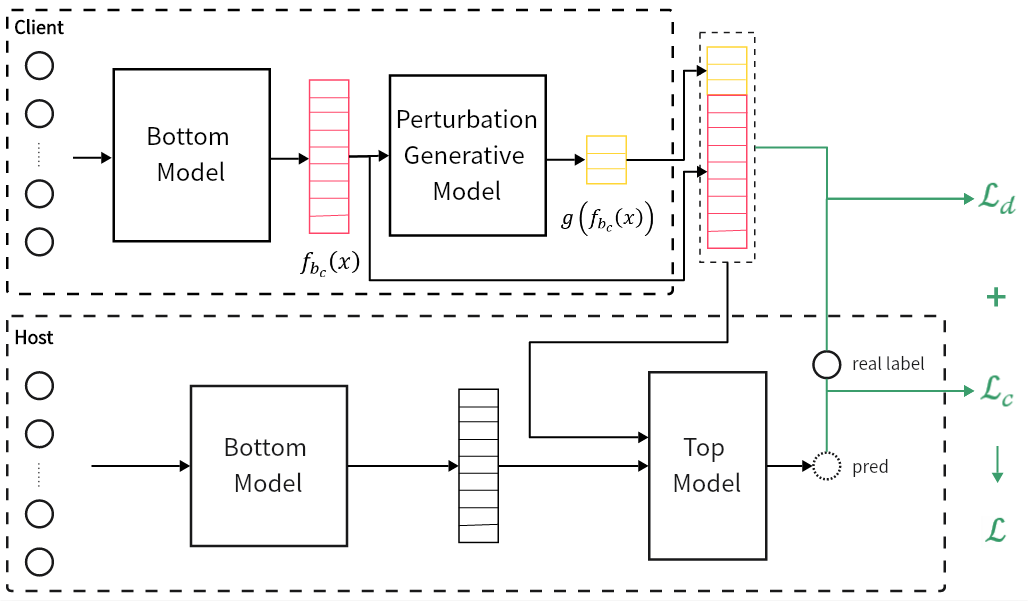}
        \caption{The workflow of the embedding extension attack. There is a perturbation generative model on the client side which takes the original embedding as input and generates the perturbation dimensions. The client appends the perturbation dimensions to the original embedding and then uploads the modified embedding to the top model.}
        \label{fig:attack}
    \end{figure}

In this section, we propose an embedding extension attack, which disrupts this defense mechanism by appending a few additional dimensions to the embedding vectors, as shown in Fig.\ref{fig:attack}. These additional dimensions are referred to as `perturbation dimensions'. According to the principle of the existing defense scheme\cite{discorloss}, if we set the perturbation dimensions to values that have extremely low or negative correlation with the label information, we can reduce the overall correlation calculation results while still maintaining a high correlation between the original embedding vectors and the label information. 

To stably disrupt this process, we design the following optimization procedure to generate perturbation dimensions that counteract the correlation of the original embedding vectors as much as possible: 
    \begin{equation}\label{eq:attack}
        \mathcal{L} = \sum_{(x,y) \in D_{aux}}Discor([g(f_{b_c}(x_c)),f_{b_c}(x_c)],y)
    \end{equation}
where $g(\cdot)$ is a simple linear neural network model held by the attacker, $x = (x_c, x_h)$ denotes a sample in the auxiliary dataset and $y$ denotes its corresponding label. Fig.\ref{fig:attack} illustrates that the perturbation generative model takes the embedding vector of the bottom model as input and generates the corresponding perturbation dimensions. The attacker trains the generative model with the labeled auxiliary data using the objective function in Eq.\ref{eq:attack}. During the training stage along with the host, the attacker feeds the embeddings into the generative model to obtain the perturbation dimensions. Subsequently, the attacker combines these perturbation dimensions with the original embedding vectors and uploads them to the top model. Alg.\ref{alg:attack} demonstrates the whole attack algorithm.

    \begin{figure}[htbp]
        \centering
        \includegraphics[width=\linewidth]{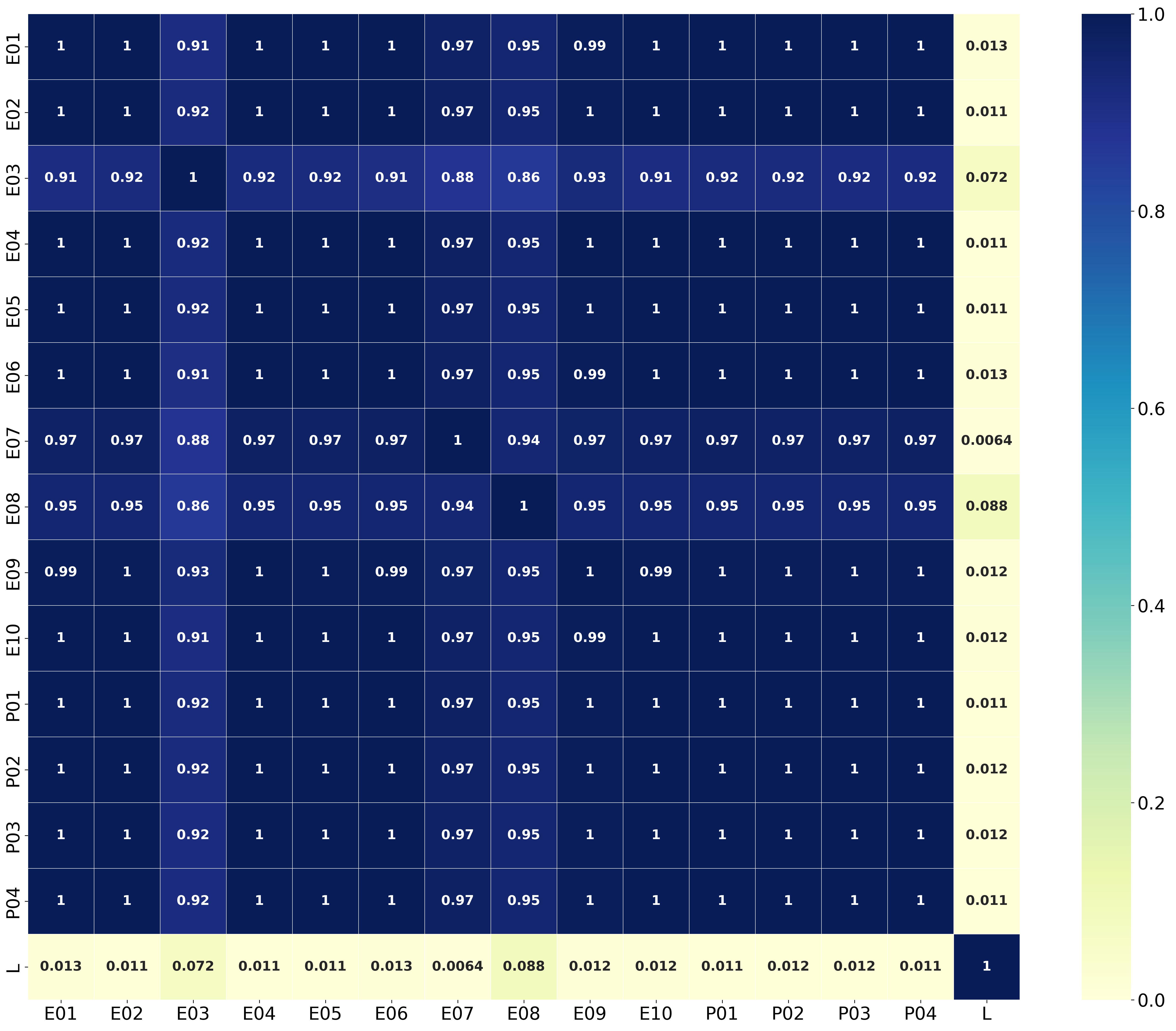}
        \caption{The Pearson correlation between the embedding dimensions and the real labels, where $P01$ to $P04$ represent the additional perturbation dimensions.}
        \label{fig:correlation}
    \end{figure}

The advantage of this approach lies in its stable disruption effect and the difficulty for the defender to counteract it (e.g., identifying or filtering out these special dimensions). This is because the defender is unaware of how much irrelevant information is contained in the uploaded embeddings. Furthermore, these special dimensions are not easy-to-filter noise. On the contrary, they exhibit correlations with other dimensions, as determined by Eq.\ref{eq:attack}. Fig.\ref{fig:correlation} also illustrates this point, showing that these perturbation dimensions exhibit similar correlation characteristics with other dimensions, without any unusual behavior. We validate the disruptive effect of this extension attack on the existing defense scheme using the Epsilon dataset, and the results indicate that under the Discorloss defense, the success rate of label inference attack significantly increases when using the extension attack with perturbation dimensions. More details can be found in Sec.\ref{eval}.

    \begin{algorithm}
        \SetAlgoLined
        \caption{Complete Embedding Extension Attack Processes in two-party SplitNN}
        \label{alg:attack}
        \KwIn{$X_{aux}$, $Y_{aux}$, $X$, $Y$, learning rate $\eta_1$, $\eta_2$}
        \KwOut{Model parameters $\theta_{b_c}$, $\theta_{g}$, $\theta_{b_h}$, $\theta_t$}
        \textbf{client} and \textbf{host} initialize $\theta_{b_c}$, $\theta_{g}$, $\theta_{b_h}$, $\theta_t$\;
        \For{$i$ in $range(EPOCH\_NUM\_1)$}{
            \textbf{client:}\\
                \For{$j$ in $range(EPOCH\_NUM\_2)$}{
                    $V_c$ = $f_{b_c}(X_{{aux}_c})$\;
                    $V_{perb}$ = $g(V_c)$\;
                    $V_c$ = $concat(V_c, V_{perb})$\;
                    $l$ = $Corr(V_c,Y_{aux})$\;
                    ${\theta_g}^{j+1}$ = ${\theta_g}^j - \eta_1\nabla_{\theta_g}{l}$\;
                }
                $V_c$ = $f_{b_c}(X_c)$\;
                $V_{perb}$ = $g(V_c)$\;
                $V_c$ = $concat(V_c, V_{perb})$\;
                upload $V_c$\;
            \ \\
            \textbf{host:}\\
                $V_h$ = $f_{b_h}(X_h)$\;
                $V$ = $concat(V_c, V_h)$\;
                $\hat{Y}$ = $f_t(V)$\;
                $l_1$ = $l_{ce}(\hat{Y},Y)$\;
                $l_2$ = $Corr(V_c,Y)$\;
                $l$ = $l_1+\lambda{l_2}$\;
            \ \\
            \textbf{client} and \textbf{host} update model parameters:\\
                ${\theta_{t}}^{i+1}$ = ${\theta_{t}}^i - \eta\frac{\partial l}{\partial \theta_{t}}$\;
                ${\theta_{b_c}}^{i+1}$ = ${\theta_{b_c}}^i - \eta\nabla_{\theta_{b_c}}l$\;
                ${\theta_{b_h}}^{i+1}$ = ${\theta_{b_h}}^i - \eta\nabla_{\theta_{b_h}}l$\;
        }
        \textbf{return}
    \end{algorithm}
 
\section{Label Obfuscation Protection Scheme}\label{defense}
The essence of the label inference attack\cite{labelinferattack} lies in clustering the embeddings at the cut layer, followed by inferring the label information corresponding to different categories from a small number of known labeled samples. To reduce its attack success rate, it's crucial to reduce the correlation between the embeddings at the cut layer and the label information, ensuring that the distribution characteristics of the embeddings significantly differ from those of the true classes.

However, as previously demonstrated, existing solutions that constrain the embeddings are prone to interference and can be compromised by attackers. Therefore, a natural approach is to design defense mechanisms based on label information, which attackers cannot access or interfere with. By disrupting the distribution of labels, the distribution of embeddings at the cut layer is naturally disrupted as well, thus reducing the correlation between the embeddings and the true labels. 

    \begin{figure}[htbp]
        \centering
        \includegraphics[width=\linewidth]{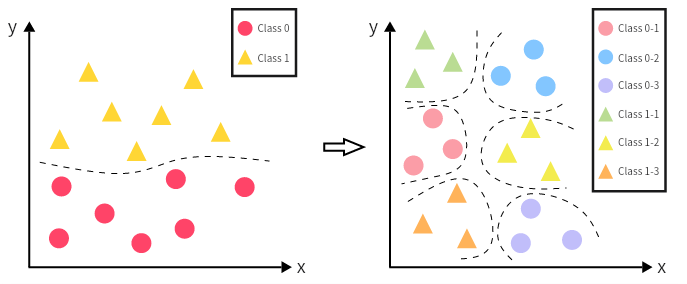}
        \caption{The key idea of LabObf is to map the original class distribution to a new scrambled distribution.}
        \label{fig:idea}
    \end{figure}

In this section, we propose \textit{LabObf}, a label obfuscation scheme that maps the original labels to a new, intentionally scrambled distribution. This mapping, through back-propagation, further disrupts the distribution at the cut layer. As shown in Fig.\ref{fig:idea}, we can artificially increase the number of categories and shuffle their numerical distributions, causing samples belonging to the same original category to spread out and reducing the distance between samples of different categories. This increases the difficulty of learning for the label inference model.

    \begin{figure}[htbp]
        \centering
        \includegraphics[width=0.9\linewidth]{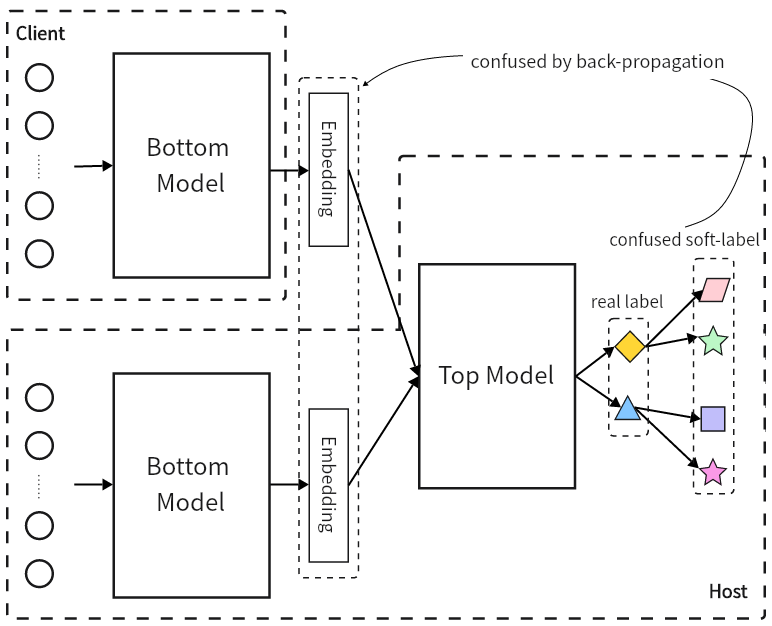}
        \caption{The workflow of LabObf. The host uses soft labels to train the VFL model. During the prediction phase, the top model generates soft labels, and the host translates them into real labels.}
        \label{fig:defense}
    \end{figure}

For example, by setting certain rules (details provided later), original samples of $0$-class are randomly assigned soft labels of three categories $[0.2, 0.6, 0.8]$, and original samples of $1$-class are randomly assigned soft labels of three categories $[0.3, 0.4, 0.9]$. Then, the new soft labels are used for training the primary task. During the prediction stage, the host infers the soft labels based on the predicted values and maps the soft labels back to the original labels. This approach reduces the implicit label information contained in the embeddings at the cut layer and prevents the leakage of true label information from back-propagated gradients. The entire training process is illustrated in Fig.\ref{fig:defense}. 

The selection of soft labels significantly impacts the effectiveness of LabObf. This can be explained in three aspects: the magnitude of soft labels, the degree of cross-over between soft labels, and the number of soft labels. We analyze these three aspects and, based on experimental experience, propose the following design principles:
\begin{enumerate}
    \item The magnitude of soft labels is significant. When decomposing the original label into soft labels, larger soft label values lead to larger gradients within the loss function, consequently slowing down the model convergence. Additionally, larger label values may introduce numerical instability during training. Conversely, decreasing the label values might cause the model to overlook subtle differences in the data, potentially leading to the model's inability to accurately capture important patterns and features, thus reducing its predictive capability. Our practical experience reveals that \textbf{normalizing the range of soft label values $R'$ to match the range of original label values $R$} is a stable method that can alleviate the training difficulties introduced by label decomposition.
    \item The degree of cross-over between soft labels is closely related to the strength of the defense. We can roughly control the defensive effect by changing the distribution of soft labels. If the soft labels of different categories do not overlap at all, for example, $0 \to [0.0, 0.4]$ and $1 \to [0.6, 1.0]$, our defense scheme would degrade to having almost no defensive effect. Conversely, if the soft labels of different categories are interleaved, the defense effect will be more ideal. Therefore, we provide an easy-to-follow design guideline for reference, i.e., for any set of $N$ soft labels randomly selected from the interval $R'$ and ordered in ascending sequence as $\{l_1, l_2, \cdots, l_N\}$, stable and effective defense can be achieved if the following two conditions are met:
    \begin{itemize}
        \item \textbf{Intervals between soft labels should not be too small}, i.e., they must satisfy: $|l_i - l_j| \ge \frac{|R'|}{2N}, \forall i,j \in \{1 \ldots N\}$. This is to prevent soft labels from being too similar to each other, which would increase the difficulty of training.
        \item \textbf{Soft labels from the same original label should not be adjacent}, i.e., they must satisfy: $Org(l_i) \ne Org(l_{i+1}), \forall i \in \{1 \ldots N-1\}$. Here, $Org(\cdot)$ denotes the function mapping a soft label to its corresponding original label. This ensures that soft labels from different original labels are appropriately interleaved.
    \end{itemize}
    \item The number of soft labels into which each original label is split also plays a crucial role in the defense's effectiveness. Having too many soft labels may increase the learning task's difficulty, requiring more complex models and additional iterations, while having too few soft labels may fail to achieve deep cross-interactions. Additionally, the number of soft labels affects the complexity and computational overhead of the defense scheme. Based on our practical experience, \textbf{splitting each original label into $2$ soft labels} is an appropriate choice. Further detailed analysis and experiments can be found in Sec.\ref{sublabel}.
\end{enumerate}

The challenge with this design is that the mapping is random, and there are no attributes in the original dataset corresponding to the post-confusion categories, making it difficult for the model to learn the distribution of soft labels. Therefore, we need to add additional attributes to the feature space to map each sample to different soft labels. The simplest approach is to add an additional attribute, with values randomly selected from $[0, 1, 2, ...]$. If the value is $0$, the sample is mapped to the first subcategory of the original category; if it is $1$, it is mapped to the second subcategory of the original category, and so on.

    \begin{figure}
        \centering
        \subfigure[Raw model.]{
              \begin{minipage}[t]{0.3\linewidth}
                  \centering
                  \includegraphics[width=\linewidth]{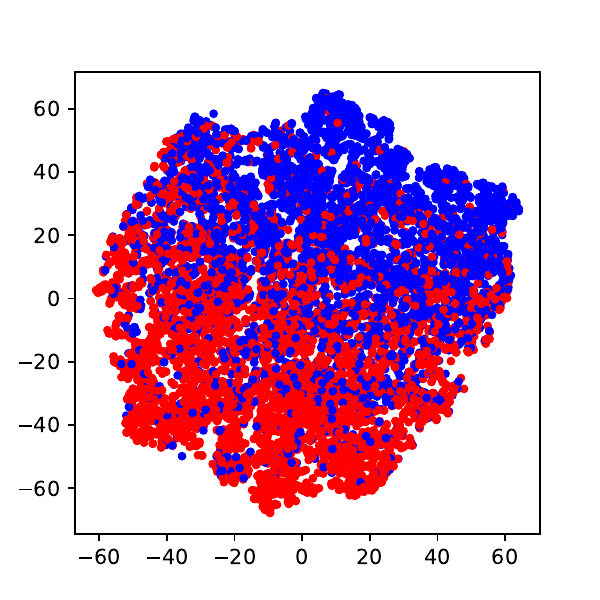}
              \end{minipage}%
        }%
        \subfigure[The client has no additional attributes.]{
              \begin{minipage}[t]{0.3\linewidth}
                  \centering
                  \includegraphics[width=\linewidth]{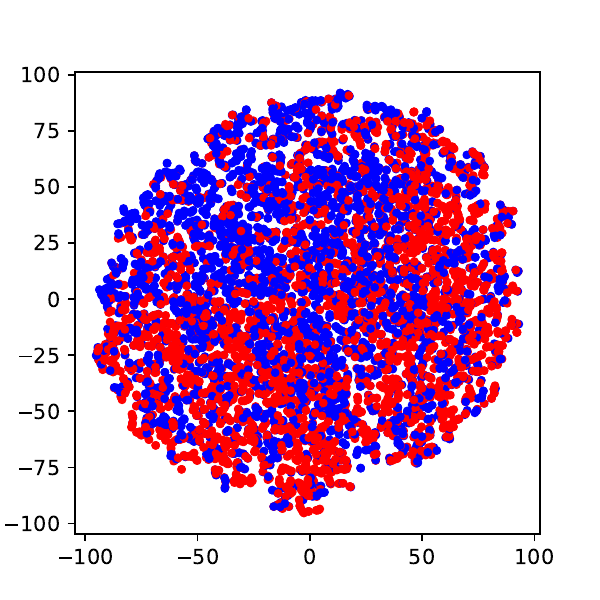}
              \end{minipage}%
        }%
        \subfigure[The client has an additional attribute.]{
              \begin{minipage}[t]{0.3\linewidth}
                  \centering
                  \includegraphics[width=\linewidth]{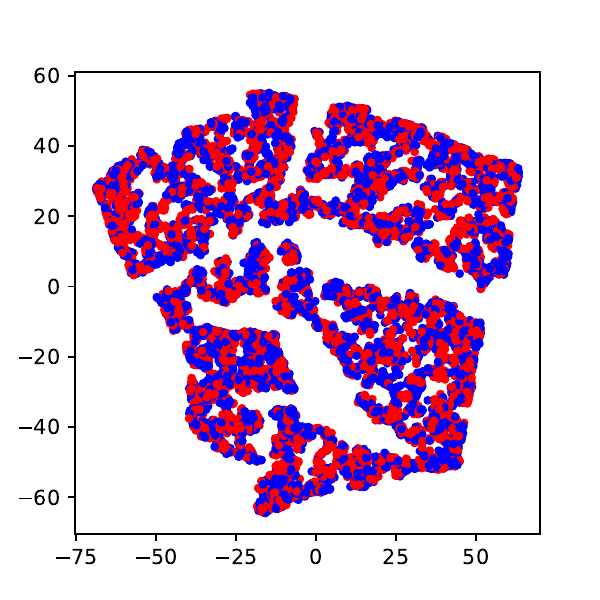}
              \end{minipage}%
        }%
        \caption{A simple example of the distributions of the client's embedding.}
        \label{fig:attribute}
    \end{figure}

Ideally, this additional attribute should only belong to the host, preventing the malicious client from inferring the mapping relationship between the true labels and the soft labels. However, in practical scenarios, we observe that when this is implemented, only the embeddings uploaded by the party holding this special attribute are disrupted, while the embeddings of the malicious client may still be close to the original classification features. As shown in Fig.\ref{fig:attribute}, when the client does not hold any additional attribute values, and only the host does, the distribution of the client's embeddings (middle image) is much closer to the distribution in the raw model (left image), indicating almost no obfuscation. However, when both the host and the client each hold an additional attribute, the distribution of the client's embeddings becomes more obfuscated and aligns more closely with the characteristics of the soft label classes (right image). Therefore, we need to design additional attributes that meet the following objectives:
\begin{itemize}
    \item They are capable of being randomly mapped to different subcategories.
    \item Their mapping relationship with the soft labels can extend to the cut layer of the malicious client.
    \item It is not possible to directly infer the mapping relationship between the original labels and the soft labels from the additional attributes held by a single participant.
    \item Even if the malicious client freely modifies the attributes it hold, it will not affect the training results.
\end{itemize}

\begin{figure}[htbp]
        \centering
        \includegraphics[width=1.0\linewidth]{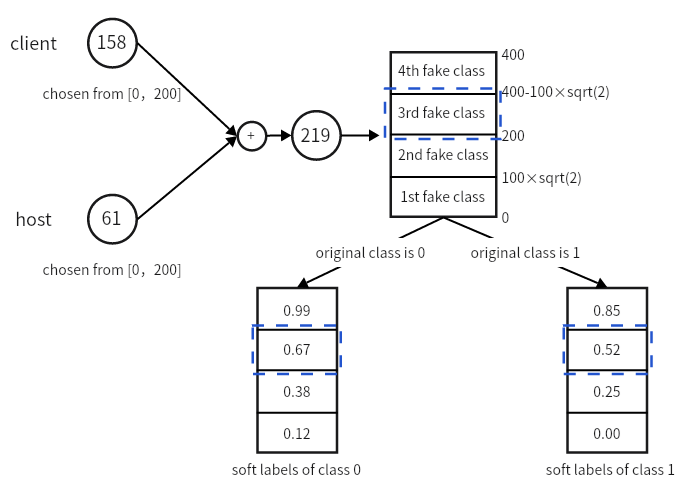}
        \caption{An example of how to build the mapping relationship between original classes and soft labels using the additional attributes.}
        \label{fig:rands}
    \end{figure}

To meet these requirements, we add two attributes, held by the host and the client respectively, with both contributing to the mapping results. In the actual experimental process, we adopt the following design: each party adds a column of special attributes, which are random numbers within a certain range. Both parties need to report their additional attribute values to the host, and the sum of these values from both parties is then used for binning to determine which subclass the samples map to, as illustrated in Fig.\ref{fig:rands}. In practice, the specific mapping relationship and binning method are determined by the host, which are unknown to the client. Therefore, it is difficult for the client to control the mapping of the obfuscated soft labels by modifying her/his own attribute value. However, if the client engages in dishonest behavior, such as reporting attribute values that do not match the actual inputs, it will lead to a decrease in the main task accuracy (from $73.84\%$ down to $50.25\%$ on the Epsilon\cite{epsilon_dataset} dataset), which is not in the client's interest.

We conduct experiments on four datasets, including both binary and multi-class datasets, covering simulated test datasets and real datasets. The results demonstrate that our scheme offers strong defense. On binary classification datasets, it can reduce the attack success rate to nearly random guessing levels, and on multi-class datasets, it can even bring the attack success rate significantly below the baseline. More details can be found in Sec.\ref{eval}.

Alg.\ref{alg:defense} illustrates the entire defense algorithm. $L_{soft}$ is a dictionary where the real labels serve as keys, each associated with an array of corresponding soft labels. $X_c['rand']$ and $X_h['rand']$ denote the additional attributes. The function $bin(\cdot)$ is used for binning, which calculates the subcategory index based on the additional attributes. $BIN\_NUM$ represents the number of subcategories for each original category.

    \begin{algorithm}
        \SetAlgoLined
        \caption{Complete LabObf Protection Processes in two-party SplitNN}
        \label{alg:defense}
        \textbf{Training Stage:}\\
        \KwIn{$X$, $Y$, $L_{soft}$, learning rate $\eta$}
        \KwOut{Model parameters $\theta_{b_c}$, $\theta_{b_h}$, $\theta_t$}
        \textbf{client} and \textbf{host} initialize $\theta_{b_c}$, $\theta_{b_h}$, $\theta_t$\;
        \ \\
        \textbf{client:}\\
            $X_c['rand']$ = $random.randint(0,200)$\;
        \ \\
        \textbf{host:}\\
            $X_h['rand']$ = $random.randint(0,200)$\;
        \ \\
        \For{$i$ in $range(EPOCH\_NUM)$}{
            \textbf{client:}\\
                $V_c$ = $f_{b_c}(X_c)$\;
                upload $V_c$, $X_c['rand']$\;
            \ \\
            \textbf{host:}\\
                $V_h$ = $f_{b_h}(X_h)$\;
                $V$ = $concat(V_c, V_h)$\;
                $\hat{Y}$ = $f_t(V)$\;
                $Y_{soft}$ = $L_{soft}[Y][bin(X_c['rand']+X_h['rand'])]$\;
                $l$ = $l_{mse}(\hat{Y},Y_{soft})$\;
            \ \\
            \textbf{client} and \textbf{host} update model parameters:\\
                ${\theta_{t}}^{i+1}$ = ${\theta_{t}}^i - \eta\frac{\partial l}{\partial \theta_{t}}$\;
                ${\theta_{b_c}}^{i+1}$ = ${\theta_{b_c}}^i - \eta\nabla_{\theta_{b_c}}l$\;
                ${\theta_{b_h}}^{i+1}$ = ${\theta_{b_h}}^i - \eta\nabla_{\theta_{b_h}}l$\;
        }
        \textbf{return}\\
        \ \\
        \hrule
        \ \\
        \textbf{Predicting Stage:}\\
        \KwIn{$x$, $L_{soft}$}
        \KwOut{$\hat{y}$}
        \textbf{client:}\\
            $v_c$ = $f_{b_c}(x_c)$\;
            upload $v_c$\;
        \ \\
        \textbf{host:}\\
            $v_h$ = $f_{b_h}(x_h)$\;
            $v$ = $concat(v_c, v_h)$\;
            $\hat{y_{soft}}$ = $f_t(v)$\;
            \For{$i$ in $range(CLASS\_NUM)$}{
                \For{$j$ in $range(BIN\_NUM)$}{
                    $dis_{min}$ = $min(dis_{min},abs(\hat{y_{soft}}-L_{soft}[i][j]))$\;
                    \If{$dis_{min}$ is updated}{
                        $\hat{y}$ = $i$\;
                    }
                }
            }
        \textbf{return} $\hat{y}$\\
    \end{algorithm}

\section{Evaluation}\label{eval}
In this section, we first describe the experimental setup, and then present the experimental results of the embedding extension attack against Discorloss\cite{discorloss}. Finally, we show the protection ability of our LabObf.




\subsection{Empirical Setup} 

    \begin{table}[htbp]
        \centering
        \caption{Detailed configuration of hardware and software.}
        \label{configdescribe}
        \scalebox{0.9}{
            \begin{tabular}{cc}
            \toprule
            \textbf{Name} &  \textbf{Configuration} \\
            \midrule
            \multirow{2}{*}{CPU} & Both of \textbf{two} are Intel(R) Xeon(R) Silver 4215R \\
                                 & CPU @ 3.20GHz with \textbf{8 cores}\\
            GPU & Both of \textbf{two} are 24GB GeForce RTX 3090\\
            Memory & 65GB\\
            OS & GNU/Linux(5.4.0-150-generic) 18.04.1-Ubuntu\\
            CUDA & the version is 11.7\\
            GCC & the version is 7.5.0\\
            Python & the version is 3.8.10\\
            Pytorch & the version is 2.1.1\\
            \bottomrule
            \end{tabular}
        }
        
    \end{table}
    
We utilize a single server with configurations outlined in Table \ref{configdescribe} to carry out all the experiments. We conduct the embedding extension attack on Epsilon\cite{epsilon_dataset} to validate the shortcomings of the existing representative defense scheme, Discorloss\cite{discorloss}. Simultaneously, we employ four diverse datasets, encompassing Epsilon\cite{epsilon_dataset}, Bank\cite{bank_marketing_222}, CovType\cite{covertype_31}, and Fault\cite{fault_dataset}, to evaluate the effectiveness of our proposed label protection scheme, LabObf. Epsilon is a standard digital test dataset featuring $2000$ attributes designed for binary classification tasks, widely utilized across the academic community. To streamline training efficiency, we randomly select a subset of $50$ features. Bank comprises genuine binary classification data, aiming to predict whether clients subscribe to a term deposit. CovType, derived from data collected by the US Geological Survey (USGS) and the US Forest Service (USFS), is a multi-classification dataset encompassing $7$ distinct forest cover types. Fault is a structured multi-classification dataset containing $3$ distinct fault types. While Epsilon and Bank serve to evaluate the efficacy of our approach in binary classification tasks, CovType and Fault are employed to explore the suitability of LabObf in multi-classification scenarios. 

The detailed experiment setups for the four datasets are presented in Table \ref{datasetdescribe}. The \textit{Features} column displays the number of features used and their partition settings. The \textit{$D_{aux}$ Size} column indicates the size of the auxiliary set of known label samples possessed by the attacker. We adopt a size for the auxiliary set similar in magnitude to that used in the label inference attack\cite{labelinferattack}. Regarding model architecture, both bottom and top models employ ReLU activated MLP as the neural network. For the experiments related to the embedding extension attack, the results presented are the average of $5$ repetitions, while for LabObf-related experiments, the results are the average of $10$ repetitions.

    \begin{table*}[htbp]
        \centering
        \caption{Detailed setups of four datasets.\\(In the \textit{Features} column, $a+b$ indicates the number of features held by the attacker and the defender.)}
        \label{datasetdescribe}
        \scalebox{0.9}{
            \begin{tabular}{c p{2.5cm}<{\centering} c p{2cm}<{\centering} c p{2cm}<{\centering} p{2cm}<{\centering}}
            \toprule
            \textbf{Dataset} &  \textbf{Train Set Size} & \textbf{Validation Set Size} & \textbf{Features} & \textbf{Number of Classes} & $\textbf{D}_{\textbf{aux}}$ \textbf{Size} & \textbf{Metric}\\
            \midrule
            Epsilon & 400000 & 100000 & 23+27 & 2 & 200 & Top-1 Acc \\
            Bank & 10000 & 1000 & 10+10 & 2 & 200 & Top-1 Acc \\
            CovType & 400000 & 180000 & 27+27 & 7 & 140 & Top-1 Acc \\
            Fault & 40000 & 10000 & 23+24 & 3 & 90 & Top-1 Acc\\
            \bottomrule
            \end{tabular}
        }
    \end{table*}

Regarding the impact of the scheme on model accuracy, we use the classification accuracy on the raw model as a baseline reference. Concerning the defensive capability of the scheme, we establish two reference metrics: 
\begin{itemize}
    \item $R_{upper}$ -- An upper bound reference value, which represents the success rate of the label inference attack on raw models and indicates the label leakage situation when no protection is applied.
    \item $R_{lower}$ -- A lower bound reference value, which represents the success rate achieved by the attacker using only the auxiliary set without accessing its bottom model for local label inference attack. This value illustrates the label leakage situation caused by attackers even when the VFL model does not disclose any label information.
\end{itemize}

\subsection{Evaluation of Embedding Extension Attack}
We validate the destructive nature of our Embedding Extension Attack against the existing mainstream defense scheme, Discorloss\cite{discorloss}, on Epsilon. In the experiments, we set the dimensions of both the attacker's and defender's bottom model output vectors to $10$, and the size of perturbation dimensions is $4$. The attacker's locally generative model $g(\cdot)$, used for Embedding Extension Attack, is a single-layer linear fully connected model with input size $10$ and output size $4$. We conduct label inference attack experiments under three settings: raw model, applying Discorloss protection without using Embedding Extension Attack, and applying Discorloss protection with Embedding Extension Attack. 

Table \ref{tab:exatk} shows the experimental results using a common hyper-parameter value (i.e., setting $\lambda$ in Eq.\ref{eq:discorloss} to $0.08$). We observe that the inclusion of Discorloss causes a slight decrease in the model's classification ability, but when the attacker uses Embedding Extension Attack to interfere with Discorloss, the model's classification ability rebounds slightly. We infer this is due to the weakening of Discorloss's constraints on the original effective dimensions of embeddings, resulting in a rebound in the primary task classification ability. In terms of defense capability, when conducting label inference attacks on the raw model without any protection, the success rate is $62.73\%$. After implementing the Discorloss protection scheme, the attack success rate drops to $51.42\%$. However, when the attacker further employs Embedding Extension Attack to interfere, the success rate of the label inference attack returns to nearly $60\%$, indicating that Discorloss can be severely disrupted by our Embedding Extension Attack.

    \begin{table}[htbp]
        \centering
        \caption{Embedding Extension Attack against Discorloss ($\lambda=0.08$) on Epsilon.}
        \label{tab:exatk}
        \scalebox{0.85}{
            \begin{tabular}{cccc}
            \toprule
              & \textbf{Raw Model} & \textbf{Discorloss w/o EA} & \textbf{Discorloss \& EA} \\
            \midrule
            \textbf{Model Acc}  & $77.77$  & $73.82$   & $74.89$            \\
            \textbf{Attack Acc} & $62.73$ & $51.42$ & $59.04$   \\
            \bottomrule
            \end{tabular}
        }
    \end{table}

\begin{figure}[htbp]
        \centering
        \includegraphics[width=\linewidth]{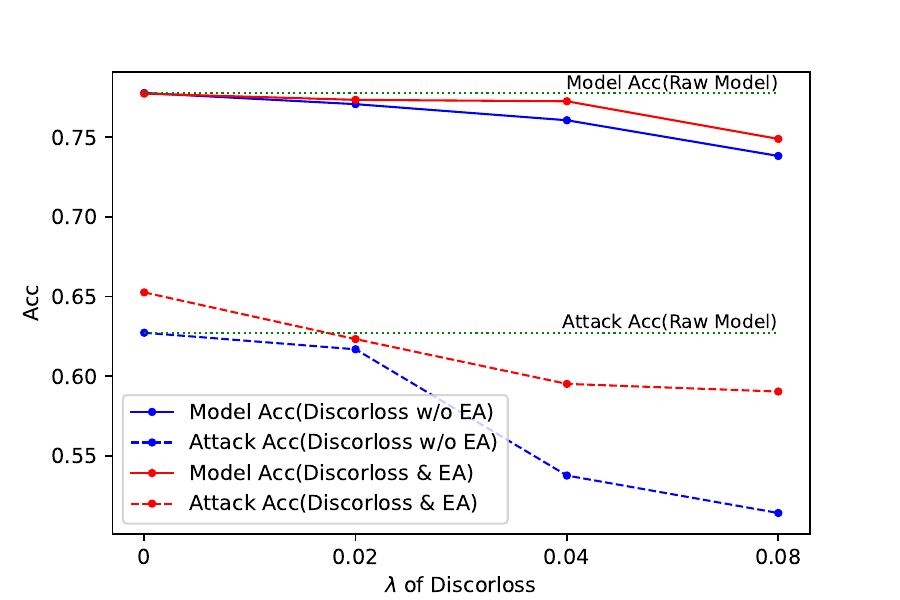}
        \caption{The effectiveness of Embedding Extension Attack varies with the hyper-parameter of Discorloss.}
        \label{fig:extenres}
    \end{figure}

Considering the significant impact of the hyper-parameter $\lambda$ on the effectiveness of Discorloss, we further conduct experiments under different hyper-parameter settings. As illustrated in Fig.\ref{fig:extenres}, when only the defender employs Discorloss for label protection, as $\lambda$ increases, the model accuracy gradually decreases, and the success rate of label inference attacks also decreases accordingly. However, when the attacker employs our proposed Embedding Extension Attack to disrupt it, we observe that regardless of the variation in $\lambda$, the model accuracy remains higher than when no disruption is applied and closer to the performance on the raw model. Additionally, the defensive capability of Discorloss is consistently severely compromised. Overall, our experimental results indicate that Discorloss can be disrupted by the attacker.

\subsection{Evaluation of LabObf}
We validate the performance of LabObf on four datasets. In the experiments, we add one-dimensional special features, $x_1$ and $x_2$, to the attacker and the defender respectively. These features have values randomly selected from the interval $[0, 200]$. Utilizing these two special features, we map each original label $l$ to a pair of soft labels $[l_1, l_2]$. The specific mapping rule is as follows: when $x_1 + x_2 \le 200$, $l$ is mapped to $l_1$; otherwise, $l$ is mapped to $l_2$. The selection of soft labels adheres to the design principles outlined in Sec.\ref{defense}. We train VFL models based on these soft labels and then applied label inference attacks to exploit predictions from the perturbed bottom model outputs.

Table \ref{tab:model_acc} shows the impact of LabObf on model accuracy. For binary classification tasks, LabObf has a negligible effect on the primary task, with accuracy loss controlled within approximately $5\%$. For multiple classification tasks, the accuracy loss slightly increases but remains within an acceptable range of about $8\%$. This increase is reasonable due to the higher number of original labels in multiple classification tasks, resulting in more mapped soft labels and greater training difficulty.

Table \ref{tab:attack_acc} demonstrates the defense performance of LabObf. We test the $R_{upper}$ and $R_{lower}$ baseline metrics on four datasets. As presented in the \textit{$R_{upper}$} row, label inference attacks show notable effectiveness across different classification tasks, with top-1 attack accuracy exceeding $60\%$ and reaching approximately $74\%$ on Fault. The \textit{$R_{lower}$} row shows the results of label inference attacks when the VFL model leaks no information, relying solely on the auxiliary set. The attack success rate exceeds $55\%$ in most cases and reaches over $69\%$ on Fault. However, with LabObf protection, the success rates of label inference attacks are significantly lower than $R_{upper}$ (dropping by more than $15\%$ at most) and even fall below $R_{lower}$ on all four datasets. This indicates that our protection scheme not only reduces label information leakage at the cut layer but also interferes with the attacker's inference. This demonstrates the effectiveness of LabObf in protecting label information from leakage.

\begin{table}[htbp]
    \centering
    \caption{The impact of LabObf on primary task accuracy across four datasets.}
    \label{tab:model_acc}
    \scalebox{0.85}{
        \begin{tabular}{ccccc}
        \toprule
             & \textbf{Epsilon}  & \textbf{Bank} & \textbf{CovType} & \textbf{Fault} \\
        \midrule
        \textbf{Raw Model} & $77.90\pm 0.23$ & $86.91\pm 0.10$ & $76.91\pm 0.02$ & $82.89\pm 0.12$ \\
        \textbf{LabObf} & $73.16\pm 0.36$ & $85.32\pm 1.31$ & $68.20\pm 1.86$ & $74.36\pm 5.98$ \\
        \bottomrule
        \end{tabular}
    }
\end{table}

\begin{table}[htbp]
    \centering
    \caption{The defense capability of LabObf against label inference attack across four datasets.}
    \label{tab:attack_acc}
    \scalebox{0.85}{
        \begin{tabular}{p{1.5cm}<{\centering}cccc}
        \toprule
           & \textbf{Epsilon} & \textbf{Bank}  & \textbf{CovType} & \textbf{Fault} \\
        \midrule
        $\textbf{R}_{\textbf{upper}}$ & $63.03\pm 3.86$ & $59.56\pm 1.75$ & $57.23\pm 0.32$ & $74.22\pm 1.34$ \\
        $\textbf{R}_{\textbf{lower}}$ & $61.27\pm 2.48$ & $55.34\pm 2.91$ & $54.03\pm 2.12$ & $69.03\pm 4.21$ \\
        \textbf{LabObf} & $50.35\pm 2.78$ & $52.44\pm 1.98$ & $43.71\pm 7.47$ & $60.17\pm 7.10$ \\
        \bottomrule
        \end{tabular}
    }
\end{table}

We evaluate the potential additional time overhead introduced by LabObf. Table \ref{traintime} reports the training times for the four datasets. Compared to the raw models, our label obfuscation scheme does increase the model training time. However, it can be observed that when the model itself has a long training time, the additional computational time introduced by LabObf is relatively minor compared to models with shorter training times. For instance, in CovType, with a training duration of $2217$s, the longest among the datasets, the additional computational time introduced by the same number of training epochs accounts for only $28.24\%$. Conversely, in Fault, with a training duration of $83$s, the shortest among the datasets, the additional computational time accounts for $248.19\%$. This discrepancy arises because the majority of the additional training time is concentrated in the preprocessing phase, which involves adding new columns to the dataset and computing soft labels. This operation has a constant time complexity and does not scale with longer training times. Therefore, considering that real-world applications often involve large datasets and complex model structures, where the training times are inherently long, the additional time overhead introduced by LabObf is acceptable in comparison.

\begin{table}[htbp]
    \centering
    \caption{Training time of LabObf across four datasets.\\(The training time is reported in the form of $a/b$, where $a$ is the total training time and $b$ is the training epochs.)}
    \label{traintime}
    \scalebox{0.9}{
        \begin{tabular}{ccc}
        \toprule
        \textbf{Training Time(s)} &  \textbf{Raw Model} & \textbf{LabObf}\\
        \midrule
        Epsilon & $(230\pm 20)$/50 & $(1277\pm 36)$/100  \\
        Bank & $(90\pm 5)$/100 & $(230\pm 13)$/300 \\
        CovType & $(2217\pm 24)$/200 & $(2843\pm 22)$/200 \\
        Fault & $(83\pm 8)$/50 & $(289\pm 8)$/50\\
        \bottomrule
        \end{tabular}
    }
\end{table}

\begin{figure}[htbp]
    \centering
    \includegraphics[width=\linewidth]{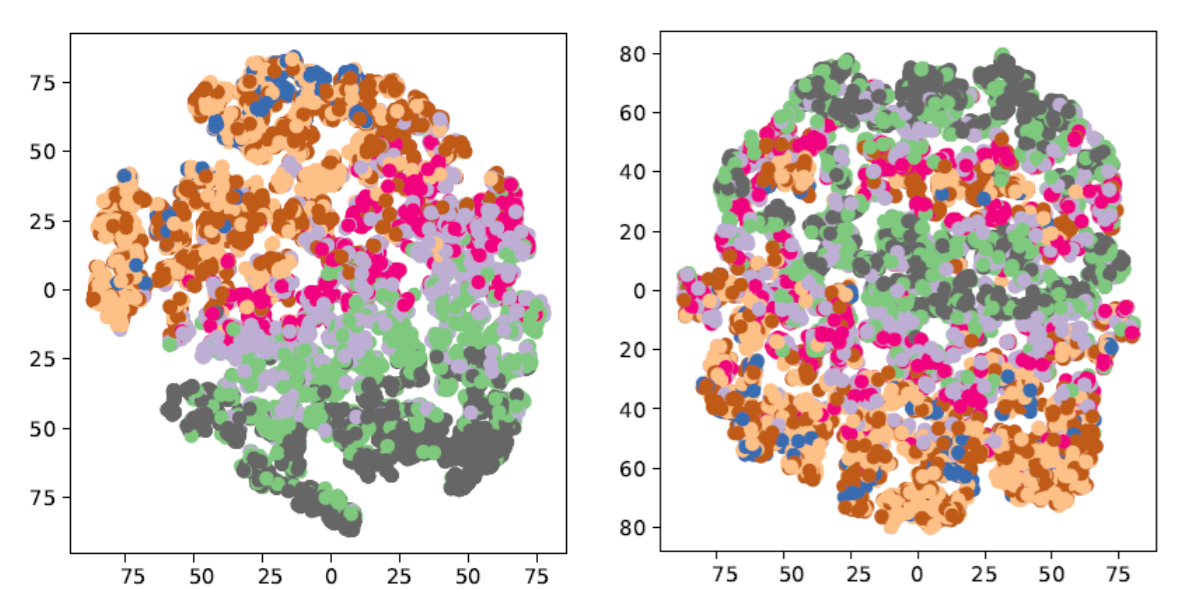}
    \caption{t-SNE visualization of attacker's bottom output on Epsilon.\\(The first plot in each column shows the distribution of clean embeddings and the second plot shows the distribution of embeddings after LabObf defense. The distributions in both plots are based on the true labels.)}
    \label{fig:tsneres}
\end{figure}

To better understand how LabObf defend against label inference attack, we use t-SNE\cite{van2008visualizing} to visualize the distribution of the attacker’s bottom model’s outputs. In  Fig.\ref{fig:tsneres}, we report the distribution of attacker's bottom output based on true labels. We can see the attacker’s bottom model learns better in separating samples from each class when there is no defense at all. LabObf mixes the distribution of various true labels in the output of the attacker's base model, thereby reducing the attacker's ability to steal labels from the trained local inference model. With LabObf, the attacker gets a trained bottom model containing less information about true labels. This is because the attacker's bottom model is trained to distinguish between soft labels, which are intermixed. Thus, after combined with the inference head, the complete model performs worse in predicting the true labels, bringing worse label inference performance. 

\begin{figure}[htbp]
    \centering
    \includegraphics[width=\linewidth]{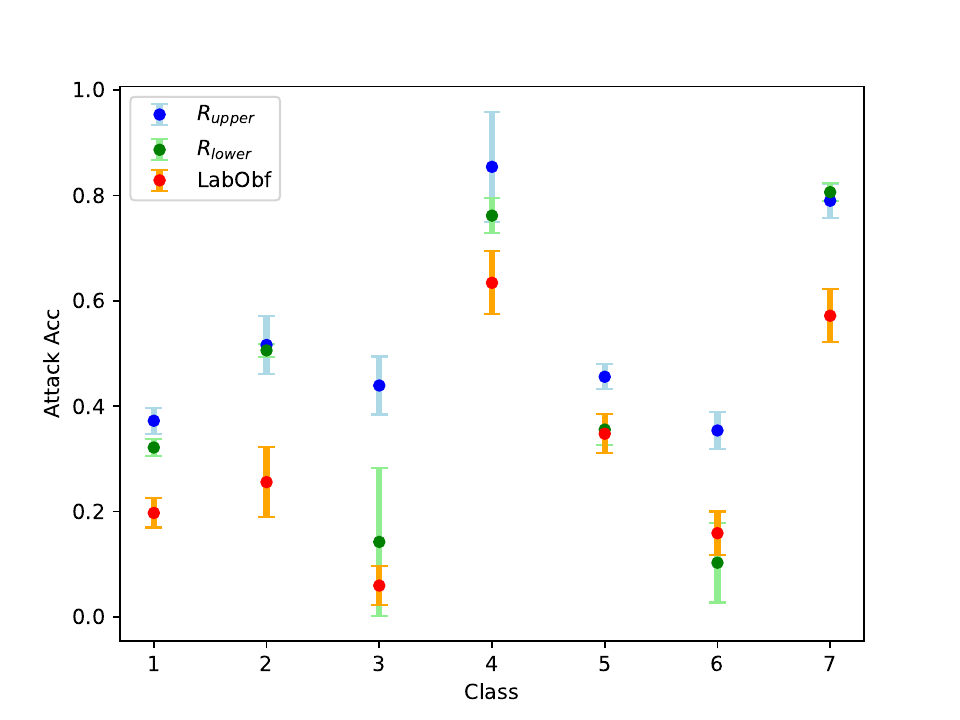}
    \caption{Attack performance on each class on CovType.}
    \label{fig:eachclass}
\end{figure}

We also analyze the protection provided by LabObf for different classes in multiple classification tasks. Fig.\ref{fig:eachclass} shows the experimental results on CovType. It can be seen that for each class, the success rate of label inference attack under LabObf protection is significantly lower than the results under $R_{upper}$. Moreover, for most classes, the success rate of attacks under LabObf protection is also much lower than under $R_{lower}$, and for the remaining classes, it is at least comparable to $R_{lower}$. This demonstrates that LabObf effectively protects each class in multiple classification tasks, ensuring no particular class is neglected.

\section{Analysis}\label{analysis}
In this section, we discuss the impact of some parameters on the defense performance of LabObf.

\subsection{Effect of Auxiliary Set Size on LabObf}
\begin{figure*}[hbp]
        \centering
        \includegraphics[width=\textwidth]{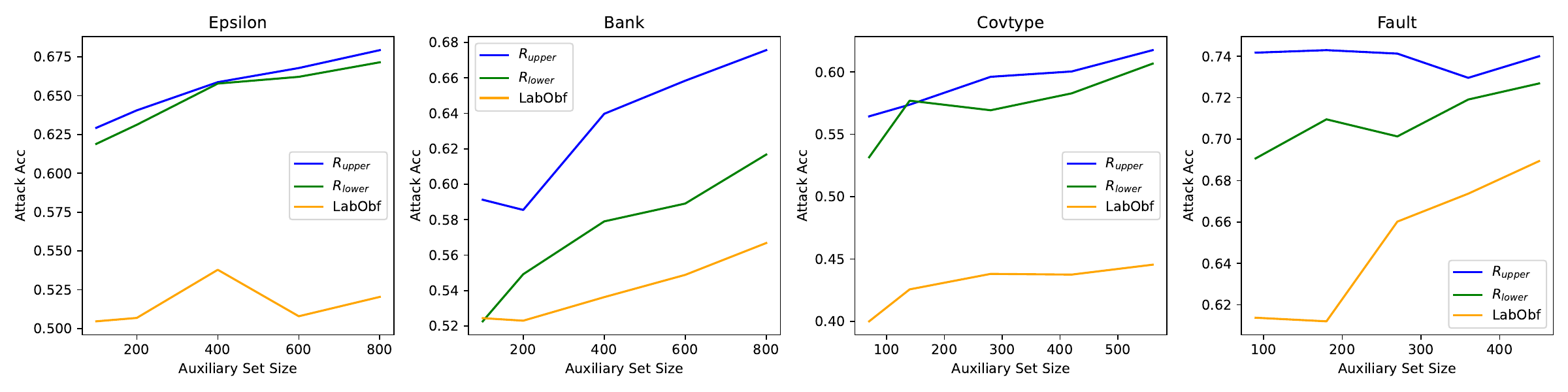}
        \caption{The label leakage under different auxiliary set size.}
        \label{fig:labelnum}
    \end{figure*}

Label inference attack utilizes additional auxiliary sets and semi-supervised learning to train local inference models. The size of the auxiliary set significantly impacts attack performance. Therefore, we further analyze the robustness of LabObf under varying auxiliary set sizes used by the attacker. Fig.\ref{fig:labelnum} shows the defense performance of LabObf across different datasets as the size of the attacker's auxiliary set varies.

For Epsilon and CovType, increasing the auxiliary set size results in a minimal improvement in label inference attacks (the $R_{upper}$ curves are relatively flat), and LabObf consistently maintains strong defense performance in these cases. In contrast, Bank is more sensitive to the size of the auxiliary set, with attack success rates increasing noticeably as the auxiliary set size grows, both in baseline settings and under LabObf. However, compared to the baseline settings, the growth rate and speed of attack success rates are slower under LabObf. For Fault, the data shows that attackers have a high success rate on the raw model, which is closer to the primary task accuracy than other datasets. Therefore, it is reasonable that increasing the auxiliary set size only slowly enhances attacks on the raw model. Both $R_{lower}$ and LabObf are more affected by the auxiliary set size. Nevertheless, the attack success rate under LabObf remains lower than in the other two cases.

Overall, as the attacker's auxiliary set size increases, the impact on LabObf also grows, and the inference attack success rate under LabObf gradually rises. This is logical. However, LabObf consistently outperforms the two baseline settings, demonstrating its superiority in protecting label privacy.

    \begin{figure*}[htbp]
        \centering
        \includegraphics[width=\textwidth]{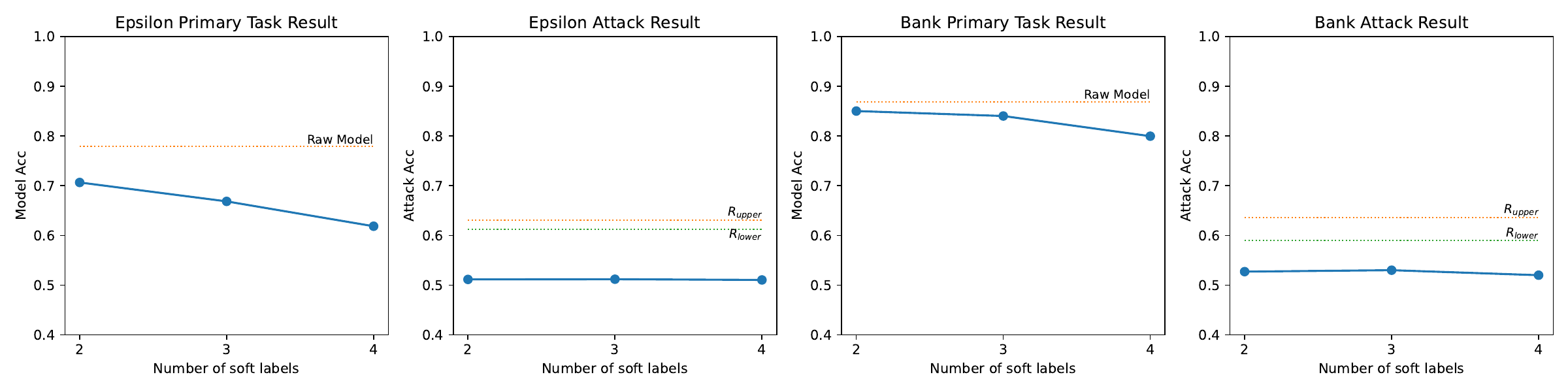}
        \caption{Model and attack performance on binary classification tasks under different soft label numbers.}
        \label{fig:sublabel}
    \end{figure*}

\subsection{Effect of Soft Label Number on LabObf}\label{sublabel}
We present a detailed analysis of how the number of soft labels influences LabObf in this section. To simplify the analysis, we assume each original label maps to an equal number of soft labels, considering three settings where each original label maps to $2$, $3$, or $4$ soft labels. Each setting is designed according to the principles for soft label magnitude and degree of cross-over in Sec.\ref{defense}. Given that the same settings on multiple classification datasets would generate too many soft labels, we experiment on the binary classification datasets Epsilon and Bank.

Fig.\ref{fig:sublabel} illustrates the model performance and attack performance under different number of soft labels. We observe that as the number of soft labels increases, the primary task accuracy decreases. This is reasonable, as an increased number of soft labels makes it harder for the model to learn and fit the soft label distribution, leading to a decrease in soft label classification accuracy, which in turn lowers the original label classification accuracy. Meanwhile, we observe that the attack success rate remains largely unchanged as the number of soft labels increases. This indicates that the model's defense performance is more influenced by the distribution of soft label values rather than their quantity. Based on these experimental results and design principles, we find that mapping each original label to $2$ soft labels allows LabObf to achieve a satisfactory balance of primary task accuracy and label privacy protection.

\section{Related Work}\label{relwork}
In the field of VFL label protection, there are other related protection schemes that warrant a simple analysis and comparison with the approach presented in this paper. \cite{complementary} designs a method called Complementary Label Coding from a knowledge distillation perspective. The main idea is to constrain the information such that the knowledge learned by the client's bottom model converges to the complement of the knowledge learned by the host's local model, aiming to reduce the information accessible to the client and thus mitigate label leakage risk. This approach focuses on optimizing the training process but lacks strong defense capabilities. If the feature data held by the malicious client is sufficiently important to achieve high prediction accuracy, this design fails to provide effective defense. Thus, even though both this scheme and our proposed label obfuscation protection scheme mention similar concepts like `label encoding', there are significant differences between the two. \cite{dispersed} designs a new dispersed training split network structure based on secret sharing, setting up a replicated but benign shadow model for the suspicious client's bottom model to share some gradients, breaking the link between the client's bottom model and the training data. However, this approach is difficult to deploy in practice. This is because the defender cannot ensure that the attacker would honestly create a benign shadow model. Moreover, placing the shadow model on the defender's side is not viable, as, under VFL rules, the defender should not know the structure of the client's bottom model. In this situation, the shadow model cannot replicate the same network structure as the client's bottom model. Label DP\cite{labeldp} is a label protection scheme based on randomized response. It randomly flips labels with a certain probability, which can obscure label information in the gradients, thereby affecting the outputs of the attacker's bottom model and making the attacker's inferences unreliable. Since randomized response simply flips labels in a rather coarse manner, it introduces additional noise that might impact the model's predictive performance, leading to significant accuracy loss. Although debiased gradient estimations are used, they cannot completely negate the effect of the label flipping. Additionally, such debiased estimations are challenging to apply to multi-class datasets. Lastly, for imbalanced datasets, the label-flipping method can introduce many false samples into the underrepresented class, impacting the accuracy of the primary task.

\section{Conclusion}\label{conclu}
In this paper, we focus on the issue of label data privacy being compromised through the leakage of embeddings at the cut layer in split learning scenarios of VFL. We first propose an embedding extension attack that could be implemented within the capabilities of a potentially malicious client acting as an attacker. By adding dimensions and manipulating the embeddings, this attack can alter the outcomes of existing defense schemes, highlighting the security risk that these defenses may be vulnerable to interference due to their reliance on data uploaded by clients. Next, we propose a label obfuscation protection scheme designed around label information that attackers cannot access or alter. This approach involves adding a small number of additional special attributes to map the original true classes to a greater number of fake classes with a more cross-referenced obfuscated value space. The primary task is then trained with these obfuscated soft labels, reducing the correlation between the cut-layer embeddings and the true labels. We validate the effectiveness of the embedding extension attack on Epsilon\cite{epsilon_dataset}, demonstrating that when the extension attack is applied, the Discorloss defense scheme\cite{discorloss} becomes ineffective, with the success rate of the label inference attack returning to levels similar to attacks on a raw model. We also evaluate the label obfuscation defense across four datasets: Epsilon\cite{epsilon_dataset}, Bank\cite{bank_marketing_222}, CovType\cite{covertype_31}, and Fault Type\cite{fault_dataset}. The results indicate that this scheme significantly reduces the success rate of the label inference attack while keeping model accuracy within an acceptable range. Our work indicates that VFL still faces risks of label leakage and introduces a novel solution aimed at fostering further development in the field of secure vertical federated collaboration.






%



\bibliographystyle{IEEEtran}
\bibliography{bib}

\begin{thebibliography}{10}
\providecommand{\url}[1]{#1}
\csname url@samestyle\endcsname
\providecommand{\newblock}{\relax}
\providecommand{\bibinfo}[2]{#2}
\providecommand{\BIBentrySTDinterwordspacing}{\spaceskip=0pt\relax}
\providecommand{\BIBentryALTinterwordstretchfactor}{4}
\providecommand{\BIBentryALTinterwordspacing}{\spaceskip=\fontdimen2\font plus
\BIBentryALTinterwordstretchfactor\fontdimen3\font minus \fontdimen4\font\relax}
\providecommand{\BIBforeignlanguage}[2]{{%
\expandafter\ifx\csname l@#1\endcsname\relax
\typeout{** WARNING: IEEEtran.bst: No hyphenation pattern has been}%
\typeout{** loaded for the language `#1'. Using the pattern for}%
\typeout{** the default language instead.}%
\else
\language=\csname l@#1\endcsname
\fi
#2}}
\providecommand{\BIBdecl}{\relax}
\BIBdecl

\bibitem{ihvfl}
F.~Tang, S.~Liang, G.~Ling, and J.~Shan, ``Ihvfl: a privacy-enhanced intention-hiding vertical federated learning framework for medical data,'' \emph{Cybersecurity}, vol.~6, no.~1, p.~37, 2023.

\bibitem{webank}
``Webank,'' \url{https://www.webank.com}, 2014.

\bibitem{IoTvfl}
A.~Fu, X.~Zhang, N.~Xiong, Y.~Gao, H.~Wang, and J.~Zhang, ``Vfl: A verifiable federated learning with privacy-preserving for big data in industrial iot,'' \emph{IEEE Transactions on Industrial Informatics}, vol.~18, no.~5, pp. 3316--3326, 2020.

\bibitem{splitnn}
P.~Vepakomma, O.~Gupta, T.~Swedish, and R.~Raskar, ``Split learning for health: Distributed deep learning without sharing raw patient data,'' \emph{ArXiv}, vol. abs/1812.00564, 2018.

\bibitem{bytedance}
``Bytedance,'' \url{https://www.bytedance.com/zh}, 2012.

\bibitem{tencent}
``Tencent,'' \url{https://www.tencent.com/zh-cn/index.html}, 1998.

\bibitem{openmined}
``Openmined,'' \url{https://www.openmined.org/}, 2017.

\bibitem{splitsec1}
Z.~Li, C.~Yan, X.~Zhang, G.~Gharibi, Z.~Yin, X.~Jiang, and B.~A. Malin, ``Split learning for distributed collaborative training of deep learning models in health informatics,'' in \emph{AMIA Annual Symposium Proceedings}, vol. 2023.\hskip 1em plus 0.5em minus 0.4em\relax American Medical Informatics Association, 2023, p. 1047.

\bibitem{splitsec2}
\BIBentryALTinterwordspacing
S.~Abuadbba, K.~Kim, M.~Kim, C.~Thapa, S.~A. Camtepe, Y.~Gao, H.~Kim, and S.~Nepal, ``Can we use split learning on 1d cnn models for privacy preserving training?'' in \emph{Proceedings of the 15th ACM Asia Conference on Computer and Communications Security}, ser. ASIA CCS '20.\hskip 1em plus 0.5em minus 0.4em\relax New York, NY, USA: Association for Computing Machinery, 2020, p. 305–318. [Online]. Available: \url{https://doi.org/10.1145/3320269.3384740}
\BIBentrySTDinterwordspacing

\bibitem{shortestpathattack}
X.~Yang, J.~Sun, Y.~Yao, J.~Xie, and C.~Wang, ``Differentially private label protection in split learning,'' \emph{arXiv preprint arXiv:2203.02073}, 2022.

\bibitem{batchattack}
T.~Zou, Y.~Liu, Y.~Kang, W.~Liu, Y.~He, Z.~Yi, Q.~Yang, and Y.-Q. Zhang, ``Defending batch-level label inference and replacement attacks in vertical federated learning,'' \emph{IEEE Transactions on Big Data}, pp. 1--12, 2022.

\bibitem{labelinferattack}
\BIBentryALTinterwordspacing
C.~Fu, X.~Zhang, S.~Ji, J.~Chen, J.~Wu, S.~Guo, J.~Zhou, A.~X. Liu, and T.~Wang, ``Label inference attacks against vertical federated learning,'' in \emph{31st USENIX Security Symposium (USENIX Security 22)}.\hskip 1em plus 0.5em minus 0.4em\relax Boston, MA: USENIX Association, Aug. 2022. [Online]. Available: \url{https://www.usenix.org/conference/usenixsecurity22/presentation/fu-chong}
\BIBentrySTDinterwordspacing

\bibitem{featureinfer}
\BIBentryALTinterwordspacing
D.~Pasquini, G.~Ateniese, and M.~Bernaschi, ``Unleashing the tiger: Inference attacks on split learning,'' in \emph{Proceedings of the 2021 ACM SIGSAC Conference on Computer and Communications Security}, ser. CCS '21.\hskip 1em plus 0.5em minus 0.4em\relax New York, NY, USA: Association for Computing Machinery, 2021, p. 2113–2129. [Online]. Available: \url{https://doi.org/10.1145/3460120.3485259}
\BIBentrySTDinterwordspacing

\bibitem{marvell}
O.~Li, J.~Sun, X.~Yang, W.~Gao, H.~Zhang, J.~Xie, V.~Smith, and C.~Wang, ``Label leakage and protection in two-party split learning,'' \emph{ArXiv}, vol. abs/2102.08504, 2021.

\bibitem{fedpass}
H.~Gu, J.~Luo, Y.~Kang, L.~Fan, and Q.~Yang, ``Fedpass: privacy-preserving vertical federated deep learning with adaptive obfuscation,'' \emph{arXiv preprint arXiv:2301.12623}, 2023.

\bibitem{flsg}
K.~Fan, J.~Hong, W.~Li, X.~Zhao, H.~Li, and Y.~Yang, ``Flsg: A novel defense strategy against inference attacks in vertical federated learning,'' \emph{IEEE Internet of Things Journal}, 2023.

\bibitem{discorloss}
\BIBentryALTinterwordspacing
J.~Sun, X.~Yang, Y.~Yao, and C.~Wang, ``Label leakage and protection from forward embedding in vertical federated learning,'' 2022. [Online]. Available: \url{https://arxiv.org/abs/2203.01451}
\BIBentrySTDinterwordspacing

\bibitem{epsilon_dataset}
``Epsilon dataset,'' [Online]. Available: \url{https://catboost.ai/en/docs/concepts/python-reference_datasets_epsilon}, 2008.

\bibitem{bank_marketing_222}
S.~Moro, P.~Rita, and P.~Cortez, ``{Bank Marketing},'' UCI Machine Learning Repository, 2012, {DOI}: https://doi.org/10.24432/C5K306.

\bibitem{covertype_31}
J.~Blackard, ``{Covertype},'' UCI Machine Learning Repository, 1998, {DOI}: https://doi.org/10.24432/C50K5N.

\bibitem{fault_dataset}
``Faulttype dataset,'' [Online]. Available: \url{https://www.kaggle.com/datasets/guanlintao/classification-of-faults-dataset}, 2023.

\bibitem{vfl}
Q.~Yang, Y.~Liu, T.~Chen, and Y.~Tong, ``Federated machine learning: Concept and applications,'' \emph{arXiv: Artificial Intelligence}, 2019.

\bibitem{b18}
\BIBentryALTinterwordspacing
O.~Li, J.~Sun, X.~Yang, W.~Gao, H.~Zhang, J.~Xie, V.~Smith, and C.~Wang, ``Label leakage and protection in two-party split learning,'' 2021. [Online]. Available: \url{https://arxiv.org/abs/2102.08504}
\BIBentrySTDinterwordspacing

\bibitem{van2008visualizing}
L.~Van~der Maaten and G.~Hinton, ``Visualizing data using t-sne.'' \emph{Journal of machine learning research}, vol.~9, no.~11, 2008.

\bibitem{complementary}
D.~Gao, S.~Wan, L.~Fan, X.~Yao, and Q.~Yang, ``Complementary knowledge distillation for robust and privacy-preserving model serving in vertical federated learning,'' in \emph{Proceedings of the AAAI Conference on Artificial Intelligence}, vol.~38, no.~18, 2024, pp. 19\,832--19\,839.

\bibitem{dispersed}
Y.~Wang, Q.~Lv, H.~Zhang, M.~Zhao, Y.~Sun, L.~Ran, and T.~Li, ``Beyond model splitting: Preventing label inference attacks in vertical federated learning with dispersed training,'' \emph{World Wide Web}, vol.~26, no.~5, pp. 2691--2707, 2023.

\bibitem{labeldp}
B.~Ghazi, N.~Golowich, R.~Kumar, P.~Manurangsi, and C.~Zhang, ``Deep learning with label differential privacy,'' \emph{Advances in neural information processing systems}, vol.~34, pp. 27\,131--27\,145, 2021.

\end{thebibliography}

\end{document}